\documentclass[11pt,a4paper]{article}
\pdfoutput=1 
\usepackage[utf8]{inputenc}
\usepackage[T1]{fontenc}
\usepackage{amsmath,amssymb}
\usepackage{graphicx}
\usepackage{booktabs}
\usepackage{array}
\usepackage{multirow}
\usepackage{makecell}
\usepackage{pdflscape}
\usepackage{caption}
\usepackage{float}
\usepackage{enumitem}
\usepackage[margin=1in]{geometry}
\usepackage{xcolor}
\usepackage{siunitx}
\usepackage{cite}
\usepackage{hyperref}
\hypersetup{
  colorlinks=true,
  linkcolor=black,
  citecolor=black,
  urlcolor=black,
  filecolor=black,
  anchorcolor=black,
  pdfborder={0 0 0}
}

\graphicspath{{./}{figures/}}

\title{\bfseries Integrating Deep Learning Demand Forecasting with
Multi-Objective Optimization for Circular Coffee Supply Chains:
A Data-Driven Framework for Cost, Emissions, and Freshness Management}

\author{%
\parbox{0.95\textwidth}{\centering
\normalsize
\mbox{Ger\c{c}ek Budak$^{a}$,}
\mbox{Faraz Gholamzadeh Gharehgheshlaghi$^{a}$,}
\mbox{Melika Barjesteh Vaezi$^{b}$,}
\mbox{Ahmad Gholizadeh Lonbar$^{c,*}$}\\[10pt]
\small\itshape
$^{a}$\,Department of Industrial Engineering, Ankara Y\i ld\i r\i m
Beyaz\i t University, Ke\c{c}i\"oren, Ankara 06010, T\"urkiye\\
$^{b}$\,Department of Kinesiology and Sport Management, Texas Tech
University, Lubbock, TX, United States\\
$^{c}$\,Department of Civil, Construction, and Environmental Engineering,
University of Alabama, Tuscaloosa, AL, USA\\
\upshape $^{*}$\,Corresponding author. E-mail:
agholizadehlonbar@crimson.ua.edu
}}

\date{}

\begin{document}
\color{black}
\maketitle

\begin{abstract}
\noindent
The coffee supply chain is one of the most complex agri-food networks in
the world, characterized by geographically dispersed production, multi-tier
stakeholder coordination, and acute sensitivity to product quality and
freshness. Although interest in sustainability and digital transformation
has grown rapidly, demand forecasting, operational optimization, and
product traceability are still typically addressed as isolated problems.
This study proposes an integrated two-phase framework that links these
traditionally separated domains. In the first phase, a hybrid Convolutional
Neural Network--Long Short-Term Memory (CNN--LSTM) architecture extracts
short-term temporal patterns through convolutional layers while capturing
long-horizon dependencies through recurrent layers. On the publicly
available \emph{Coffee Chain Sales} dataset, and using a strictly
chronological 70/15/15 train/validation/test partition, the model attains a
mean absolute error (MAE) of $22.87$ and a coefficient of determination
($R^2$) of $0.90$, improving on the strongest standalone deep-learning
benchmark by roughly $12\%$ and on classical statistical methods by more
than $30\%$. In the second phase, the resulting time-indexed demand matrix
is embedded directly into a tri-objective mixed-integer linear programming
(MILP) model that simultaneously minimizes total economic cost, minimizes
carbon emissions, and maximizes delivered freshness across a multi-period,
multimodal, and closed-loop network with primary and backup suppliers and
circular waste recovery. Freshness is modeled as an exponential decay
function of inventory age, so that operational decisions directly shape
delivered quality. The $\varepsilon$-constraint method produces $25$
Pareto-efficient solutions, and single- and bivariate sensitivity analyses
reveal pronounced nonlinear threshold effects in demand, carbon price, and
maximum allowable product age. Policy-scenario analysis shows that a
balanced sustainability policy cuts emissions by $22.4\%$ for only a $9.9\%$
cost increase while keeping freshness within $1.7\%$ of the baseline,
confirming that coordinated, moderate policies outperform extreme
single-objective strategies.

\end{abstract}

\section{Introduction}\label{sec:intro}

The global coffee supply chain ranks among the most extensive and complex
commodity networks in the world economy. It begins with smallholder farms
across tropical regions in Latin America, Africa, and Southeast Asia and
extends through layers of processing stations, local collectors, exporters,
multimodal transportation networks, importers, roasters, distributors, and
retailers before finally reaching end consumers. This chain encompasses a
long sequence of activities including harvesting, primary processing,
drying, grading, transportation to collection hubs, secondary processing,
international logistics, roasting, and packaging. Major producers such as
Brazil, Vietnam, Colombia, Indonesia, and Ethiopia supply a substantial
share of global demand, and their products travel long distances through
multimodal logistics systems before reaching markets in North America,
Europe, East Asia, and the Middle East. The geographic dispersion and the
diversity of actors make coffee a focal topic in the supply chain
literature, with significant economic, social, and environmental
implications.

Beyond its structural complexity, the coffee supply chain is highly
sensitive to climatic conditions, global price volatility, quality
variation, and regulatory standards. Climate change can disrupt production,
while fluctuations in commodity markets and the dynamics of international
trade shape the strategic behavior of exporters and importers. Quality,
which depends on harvesting methods, drying processes, and roasting
conditions, plays a decisive role in the final value of coffee. As a
result, stakeholders across the chain must make decisions under uncertainty
regarding demand, prices, product freshness, and environmental constraints.

\subsection{The Multimodal Coffee Supply Chain}
Because coffee is transported over long distances and passes through
multiple intermediate layers, relying on a single transportation mode is
economically inefficient and operationally risky; the coffee industry is
therefore intrinsically multimodal. Early stages typically rely on road
transport for flexibility over short distances. Movement from aggregation
centers to ports often combines road and rail, while international shipments
are conducted mainly via containerized maritime transport owing to its cost
efficiency and capacity. In special cases---high-value micro-lots, urgent
deliveries, or premium specialty coffee---air transport ensures rapid
delivery and quality preservation. Each mode carries distinct constraints:
variable lead times, limited freight capacity, port congestion, customs
delays, volatile shipping rates, and differing environmental footprints.
Any misalignment among these layers can lengthen cycle times, reduce
freshness, raise logistics costs, and degrade service levels.

\subsection{Supply Chain Traceability}
Traceability is a foundational capability in modern agri-food value chains
and is especially critical in coffee due to the extensive geographical
dispersion of production, processing, and consumption. At every stage,
product attributes such as origin, quality grade, processing method, and
sustainability certifications must be preserved with accuracy. Growing
demand for ethically sourced and environmentally responsible products has
intensified the importance of traceability: consumers, regulators, and
certification bodies expect transparent documentation of origin, labor
practices, environmental impact, and quality assurance. Technologies such
as blockchain, RFID, barcodes, QR codes, and cloud-based monitoring enable
real-time visibility, fraud prevention, and cross-border compliance, but
their effectiveness depends on standardized data protocols, digital
infrastructure at origin, and collaboration among participants. In this
study, traceability is conceptualized as a data-driven layer linking
physical flows with digitally recorded attributes and is embedded in the
optimization model through freshness indices, age-evolution equations,
transportation lead times, and circular-flow records.

\subsection{Data-Driven Optimization}
Data-driven optimization is a paradigm in which planning decisions are
guided by empirical data rather than fixed assumptions. In complex supply
chains, where uncertainties in demand, prices, yields, lead times, and
quality are prominent, relying solely on deterministic averages yields
suboptimal decisions. Data-driven approaches incorporate historical
records, real-time information, and predictive analytics into the
optimization layer, capturing temporal patterns, seasonality, and
structural dependencies that traditional frameworks overlook. In the
present research, data-driven optimization is the mechanism that links
machine-learning forecasting to the tri-objective mathematical model: the
forecasting engine produces time-dependent, product-specific demand
estimates that are inserted directly into the optimization model through the
demand constraints.

\subsection{Problem Statement, Contributions, and Organization}
Despite a decade of research attention, work on coffee supply chains remains
fragmented. Much of it targets a single dimension---sustainability
assessment, governance, certification, or traceability technology---without
integrating these into a unified analytical framework for operational
decision-making. Many models rely on qualitative analysis or
single-objective optimization, and most treat demand as deterministic and
static, ignoring the nonlinear temporal patterns, seasonality, and
cross-channel heterogeneity of real markets. A particularly significant gap
is the limited use of data-driven, predictive approaches that link demand
forecasting directly to multi-objective, traceability-aware planning.

To address these gaps, this study develops an integrated two-phase framework
that links data analytics with optimization-based decision-making. The
overall objective is the application of machine-learning methods for
optimization and traceability in the coffee supply chain. The associated
sub-objectives are: (i) to develop a deep-learning forecasting model that
generates multi-product demand predictions across sales channels; (ii) to
integrate the forecasting outputs into a multi-objective model for tactical
planning; (iii) to formulate a tri-objective model that simultaneously
minimizes cost, minimizes emissions, and maximizes freshness; and (iv) to
design traceability mechanisms across the supply chain. Accordingly, the
study is guided by four research questions concerning predictive accuracy,
forecast--optimization integration, the trade-offs among the three
objectives, and the design of a reliable traceability mechanism.

The primary contributions are fourfold. \textbf{(i)} We explicitly integrate
data-driven forecasting with multi-objective optimization, so that planning
decisions rest on realistic, time-dependent demand patterns rather than
static assumptions. \textbf{(ii)} We model freshness as an independent
optimization objective with dynamic age tracking across echelons, giving a
faithful representation of quality--time trade-offs absent from most
existing models. \textbf{(iii)} We develop a comprehensive closed-loop model
capturing multimodal transportation, primary/backup supplier resilience, and
circular waste recovery in a single formulation. \textbf{(iv)} We conduct
extensive sensitivity and policy-scenario analyses that reveal nonlinear
threshold effects and yield actionable managerial insight. The remainder of
the paper is organized as follows. Section~\ref{sec:lit} reviews the
literature; Section~\ref{sec:method} presents the methodology;
Section~\ref{sec:results} reports computational results; and
Section~\ref{sec:conc} concludes.

\section{Literature Review}\label{sec:lit}

\subsection{Coffee Supply Chain Management}
The coffee supply chain has attracted considerable attention because of its
economic significance, social implications, and environmental challenges.
Kittichotsatsawat et al.~\cite{Kittichotsatsawat2024} designed
sustainability-assessment indicators for Thailand's coffee supply chain
using Axiomatic Design and the Business Model Canvas, providing a framework
for multi-dimensional performance evaluation. Abdullah et
al.~\cite{Abdullah2025} examined governance structures and supply-chain
finance in Indonesia's specialty-coffee industry, finding that cooperative
models reduce transaction costs and enhance financial inclusion for
smallholders.

Blockchain has emerged as a prominent theme. G\'omez and
Garbinato~\cite{Gomez2025} conducted a systematic review of blockchain for
coffee traceability, identifying persistent barriers including the absence
of technical standards, limited digital infrastructure at origin, and poor
interoperability. Rahman et al.~\cite{Rahman2025} evaluated
blockchain-enabled traceability in African coffee value chains, documenting
gains in transparency and farmer premiums alongside adoption barriers.

From an optimization perspective, Zohourfazeli et
al.~\cite{Zohourfazeli2025} developed a MILP model for circular
coffee-waste networks, optimizing location, allocation, and routing for
spent-coffee-grounds recovery; Ch\'avez et al.~\cite{Chavez2018} optimized
a biofuel supply chain from coffee crop residues; and Clavijo-Buritica et
al.~\cite{Clavijo2022} addressed multi-period production planning in
Colombian coffee chains under demand uncertainty. Systematic reviews
\cite{DeFelice2025} confirm that the literature remains fragmented, with
most models limited to cost minimization or environmental assessment
without jointly addressing cost, emissions, and freshness.

\subsection{Data-Driven Approaches in Supply Chains and Related Domains}
Machine learning and deep learning are now central to supply-chain
decision-making. Razmi et al.~\cite{Razmi2025} reviewed ML for biomass
supply chains and noted that integration of ML with multi-objective
optimization remains limited; Gabellini et al.~\cite{Gabellini2025} proposed
a cost-aware ML framework for logistics demand forecasting with asymmetric
loss. Hybrid architectures that combine complementary neural components have
proven especially effective for complex temporal patterns: Taghiyeh et
al.~\cite{Taghiyeh2023} introduced a multi-phase hierarchical forecasting
approach for retail supply chains, and Livieris et al.~\cite{Livieris2020}
showed that CNN--LSTM hybrids outperform standalone architectures, with
convolutional layers extracting local features and LSTM layers capturing
long-range dependencies.

The methodological value of \emph{hybrid and combined neural architectures}
is corroborated well beyond supply-chain settings. In flood-inundation
mapping, Seyvani et al.~\cite{SeyvaniCNNFC2024} proposed a novel algorithm
combining a CNN with fully connected networks, and related work integrated
heterogeneous physical models through a neural
surrogate~\cite{NikrouNNint2025}; sub-matrix convolutional designs have been
used for benchmark mapping from aerial imagery~\cite{SeyvaniSubmatrix2025},
while transferable deep-learning models trained on simulation outputs and
synthetic hydrographs have demonstrated cross-domain
generalization~\cite{NikrouTransferable2025}. These studies reinforce a
central premise of the present work: pairing a local-feature extractor with
a sequence model yields more robust predictors than either component alone.
In a similar spirit, hybrid physics-informed neural networks combined with
digital-twin and blockchain layers have been used to optimize energy
consumption in smart buildings~\cite{GholizadehPINNDT2025}, illustrating how
hybrid learning architectures couple naturally with optimization and secure
data layers.

A second relevant thread is \emph{ML as a decision-support engine}. Nikrou
et al.~\cite{NikrouDecision2025} demonstrated machine learning for timely
operational decision-making, and explainable ML has been used to predict
correction factors that improve operational
frameworks~\cite{BaruahXAI2025}. Data-driven decision support has likewise
been applied to sustainable-welfare policy analysis in the presence of
nonlinear \emph{threshold effects}~\cite{GholizadehThreshold2024}; this is
directly analogous to the threshold-driven responses we observe in
Section~\ref{sec:results}, where cost, emissions, and freshness shift
abruptly once binding constraints are reached.

A third thread concerns \emph{digital infrastructure and AI adoption} that
underpins traceability. AI-enabled digital twins have been deployed for
large-scale infrastructure monitoring in smart-city
settings~\cite{GholizadehDT2024}, IoT frameworks have enabled intelligent
analysis and energy management across smart cities~\cite{GholizadehIoT2025},
and foundation-model segmentation has been applied to automated asset/defect
assessment~\cite{AhmadiSAM2025}---all examples of the sensing-plus-analytics
stack on which supply-chain traceability ultimately depends. Finally,
because traceability succeeds only when end users trust and adopt the
technology, human-centered work on AI-driven customer
service~\cite{EsmaeiliAICS2024} and on user adoption of digital tools in
industry settings~\cite{MirzaeiBIM2025} is pertinent to the consumer-trust
and transparency goals motivating this study.

\subsection{Freshness Management and Traceability}
Freshness-aware planning has become a specialized stream for perishables. Yu
et al.~\cite{Yu2024} developed integrated cold-chain planning with
temperature-dependent quality decay; Yuan et al.~\cite{Yuan2024} proposed
time-sensitive routing for fresh-food distribution; and Zhang et
al.~\cite{Zhang2024} analyzed freshness-keeping effort and value-added
service choices, showing that coordination mechanisms shape equilibrium
quality. On the traceability side, Sreenivasan and
Suresh~\cite{Sreenivasan2024} examined IoT-enabled visibility in cold-chain
logistics, and Liu et al.~\cite{Liu2025} proposed dynamic optimization for
e-commerce fresh-product chains with blockchain integration and reference
effects. Complementing these operational studies, IoT/analytics
stacks~\cite{GholizadehIoT2025}, blockchain-secured data
layers~\cite{GholizadehPINNDT2025}, and AI-mediated customer
interfaces~\cite{EsmaeiliAICS2024} indicate how real-time traceability data
can ultimately be coupled to optimization and to consumer trust.

\subsection{Scientometric Overview and Research Gap}
To position the present study, a scientometric analysis was conducted with
VOSviewer over three domains---\emph{coffee supply chain}, \emph{data-driven
models}, and \emph{freshness control and traceability}---using co-occurrence
of keywords, country collaboration, and co-authorship networks. The
keyword analysis (Table~\ref{tab:scientometric}) shows that the coffee
domain is organized around management/optimization, circular economy,
environmental sustainability, digitalization/traceability, and
certification; the data-driven domain clusters around machine learning,
mathematical modeling, and adaptive data-based control; and the freshness
domain clusters around quality/shelf-life, intelligent packaging, and
IoT/sensor monitoring. Synthesizing these strands reveals a persistent gap:
no existing study unifies deep-learning forecasting, digital-traceability
considerations, and multi-objective coffee supply chain optimization within
a single, coherent, data-driven decision-support framework. The present
study is designed to fill precisely this gap.

\begin{table}[H]
\centering
\caption{Top keywords by occurrence and total link strength (TLS) across
the three scientometric domains.}
\label{tab:scientometric}
\begin{tabular}{llcc}
\toprule
\textbf{Research domain} & \textbf{Keyword} & \textbf{Occurrences} &
\textbf{TLS}\\
\midrule
\multirow{3}{*}{Coffee supply chain}
 & coffee          & 111 & 687\\
 & certification   & 72  & 539\\
 & sustainability  & 75  & 482\\
\midrule
\multirow{3}{*}{Data-driven models}
 & mathematical models & 111 & 505\\
 & data models         & 57  & 289\\
 & data-driven         & 76  & 177\\
\midrule
\multirow{3}{*}{\makecell[l]{Freshness control\\and traceability}}
 & quality   & 104 & 273\\
 & freshness & 105 & 204\\
 & chitosan  & 33  & 130\\
\bottomrule
\end{tabular}
\end{table}

\section{Methodology}\label{sec:method}

\begin{figure}[H]
\centering
\includegraphics[width=0.9\textwidth]{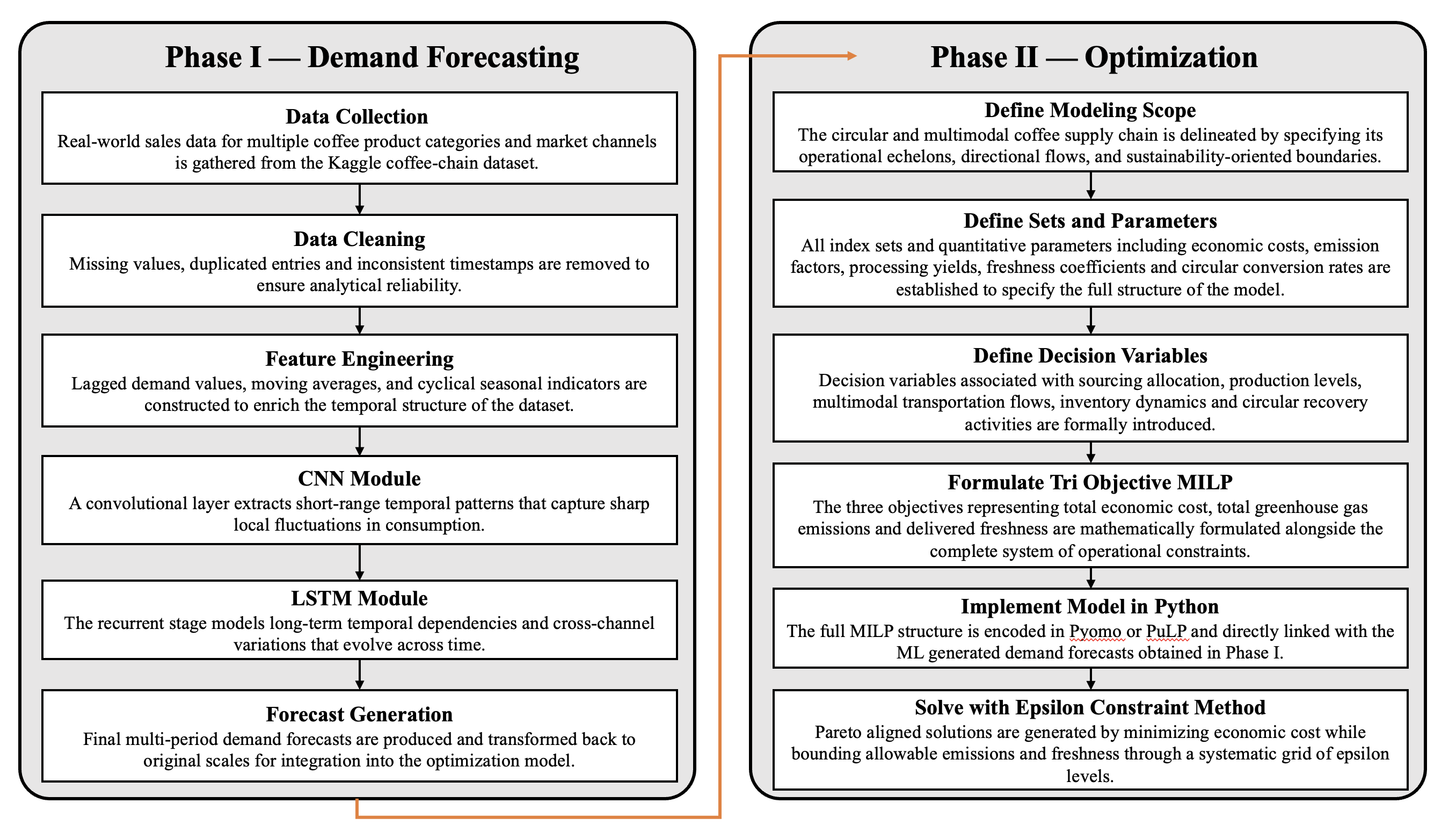}
\caption{Integrated two-phase framework: demand forecasting (Phase~I)
feeding multi-objective optimization (Phase~II).}
\label{fig:overall}
\end{figure}

The framework comprises two integrated phases: a deep-learning demand
forecasting module (Phase~I) and a tri-objective optimization model
(Phase~II). The forecasts generated in Phase~I are passed, without manual
intervention, as time-indexed demand parameters to Phase~II. The overall
structure is shown in Figure~\ref{fig:overall}.

\subsection{Supply Chain Network Structure}
\begin{figure}[H]
\centering
\includegraphics[width=0.9\textwidth]{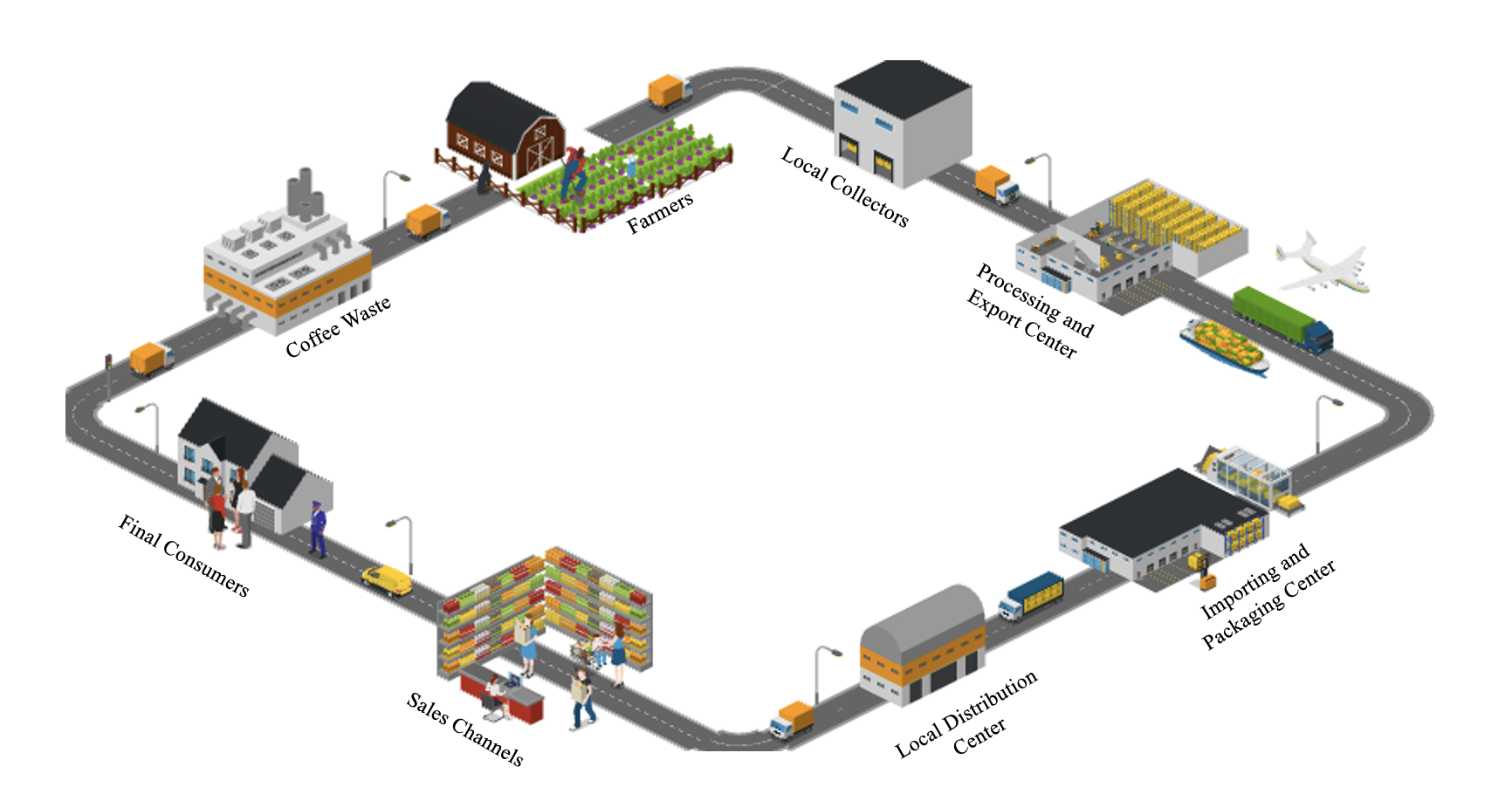}
\caption{Circular, multimodal coffee supply chain network with forward and
reverse material flows.}
\label{fig:network}
\end{figure}

The network (Figure~\ref{fig:network}) spans primary suppliers ($S^{P}$) and
backup suppliers ($S^{B}$); processing/roasting plants ($P$); distribution
centers ($D$); retail/market nodes ($R$); and circular processing
facilities ($F$). At the upstream level, green coffee is procured from a
heterogeneous portfolio of primary suppliers (stable, contractual, high
quality) and backup suppliers (activated under price shocks, weather-induced
shortages, or disruptions, enhancing resilience). At the midstream level,
processing and roasting facilities convert green coffee into finished
products. Downstream, distribution centers hold inventory and allocate
finished products to heterogeneous market channels (retail, caf\'es,
restaurants, HoReCa). The circular subsystem collects spent grounds from
markets and delivers them to circular facilities, whose recovered outputs
re-enter the chain at suppliers (biological loop) or plants (industrial
loop). Transportation between all layers uses maritime, road, and air
modes, each with distinct cost, lead time, capacity, and emission intensity.

\subsection{Phase I: Demand Forecasting}

\subsubsection{Data Sources and Selection}
Several data sources are commonly used in coffee supply chain research, each
covering a different segment of the chain. Country-level production data
(e.g., FAOSTAT) and industry trade/price statistics (e.g., the
International Coffee Organization) provide a macro view well suited to
long-term trend and market-dynamics analysis but are too aggregated for
operational, transaction-level modeling. Standardized trade datasets (e.g.,
ITC Trade Map) support export-market analysis, and report-based outlooks
(e.g., USDA-FAS) support contextualization and validation. In contrast,
Kaggle frequently provides micro-level, decision-oriented sales data that
align with demand modeling, temporal pattern analysis, and operational
decision-making. Table~\ref{tab:datasets} summarizes these sources. On this
basis, the \emph{Coffee Chain Sales} dataset is selected as the empirical
basis for the forecasting layer.

\begin{landscape}
\begin{table}[H]
\centering
\caption{Summary of data sources commonly used in coffee supply chain
research.}
\label{tab:datasets}
\footnotesize
\setlength{\tabcolsep}{4pt}
\begin{tabular}{|>{\raggedright\arraybackslash}p{24mm}|
>{\raggedright\arraybackslash}p{26mm}|
>{\raggedright\arraybackslash}p{26mm}|
>{\raggedright\arraybackslash}p{52mm}|
>{\raggedright\arraybackslash}p{48mm}|
>{\raggedright\arraybackslash}p{40mm}|}
\hline
\textbf{Source} & \textbf{Level} & \textbf{Type} & \textbf{Key features} &
\textbf{Primary use} & \textbf{Limitations}\\ \hline
FAOSTAT / UNData & Country, annual & Production & Production volume,
harvested area, yield, country/region & Macro supply trends, cross-country
comparison & No demand, price, inventory, or operational variables\\ \hline
ICO & Country/region, monthly--annual & Trade \& price & Exports, imports,
consumption, inventories, price indices & Market dynamics, price behavior,
industry time series & High aggregation; not transactional\\ \hline
ITC Trade Map & Country pairs, annual & Trade & Trade value/volume by HS
code, market share, growth & Export-market and competitiveness analysis &
No sales, demand, or operational variables\\ \hline
USDA (FAS) & Country, annual & Outlook & Production, consumption, ending
stocks, expert assessment & Trend validation, parameter calibration &
Report-based; not for direct ML training\\ \hline
Kaggle -- Shop Sales & Store, daily/transactional & Sales & Date/time,
store, product, quantity, price, total & Demand forecasting, temporal
patterns, replenishment & Specific stores; not industry-representative\\
\hline
Kaggle -- Chain Sales & Branch/market, periodic & Sales \& finance & Sales,
profit, cost, marketing, inventory, targets & Managerial decision support,
optimization, scenarios & Partial coverage; limited scope\\ \hline
\end{tabular}
\end{table}
\end{landscape}

\subsubsection{Preprocessing Pipeline}
The \emph{Coffee Chain Sales} dataset comprises $1{,}062$ records and $21$
variables capturing sales, cost, profit, inventory, and market attributes
across U.S. coffee retail channels, with \texttt{Sales} taken as the demand
signal. Preprocessing proceeds in five steps. \textbf{(1)} Missing
observations (approximately $3.2\%$ of records) are imputed by linear
interpolation between the nearest valid neighbors,
\begin{equation}
y_t=\tfrac{1}{2}\!\left(y_{t^-}+y_{t^+}\right),
\end{equation}
preserving temporal continuity. \textbf{(2)} Outliers are detected with a
rolling window of length $w=14$ and threshold $\kappa=3$: an observation is
flagged when
\begin{equation}
\lvert y_t-\mu_t^{w}\rvert>\kappa\,\sigma_t^{w},
\end{equation}
and replaced by the local mean $\mu_t^{w}$, where $\mu_t^{w}$ and
$\sigma_t^{w}$ are the rolling mean and standard deviation. \textbf{(3)}
Each series is min--max normalized to $[0,1]$,
\begin{equation}
y_t=\frac{y_t-y_{\min}}{y_{\max}-y_{\min}}.
\end{equation}
\textbf{(4)} Supervised samples use a lookback window of $L=30$,
\begin{equation}
\mathbf{x}_t=(y_{t-1},y_{t-2},\dots,y_{t-L}),
\end{equation}
augmented with cyclical calendar encodings $\sin(2\pi m_t/12)$ and
$\cos(2\pi m_t/12)$ to represent seasonality smoothly. \textbf{(5)} The
series is partitioned \emph{chronologically} into $70\%$ training, $15\%$
validation, and $15\%$ test, with \emph{no} shuffling so that evaluation
occurs strictly on unseen future periods:
\begin{equation}
T_{\text{train}}\prec T_{\text{val}}\prec T_{\text{test}},\qquad
T_{\text{train}}\cup T_{\text{val}}\cup T_{\text{test}}=T,\qquad
\text{pairwise disjoint.}
\end{equation}
This single chronological 70/15/15 protocol is used throughout the study.

\subsubsection{Hybrid CNN--LSTM Architecture}
\begin{figure}[H]
\centering
\includegraphics[width=0.85\textwidth]{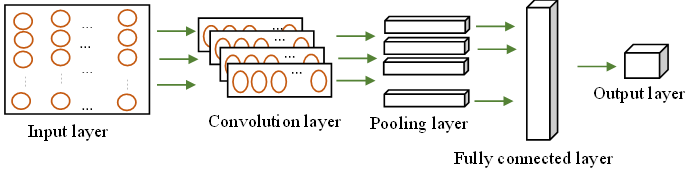}
\caption{CNN feature-extraction block for short-range temporal patterns.}
\label{fig:cnn}
\end{figure}

\begin{figure}[H]
\centering
\includegraphics[width=0.85\textwidth]{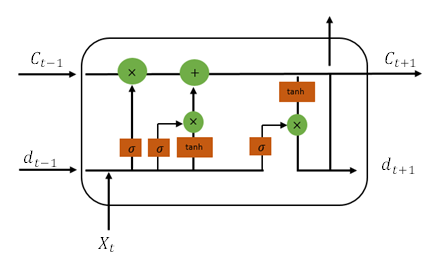}
\caption{LSTM cell structure for long-horizon temporal dependencies.}
\label{fig:lstm}
\end{figure}

\begin{figure}[H]
\centering
\includegraphics[width=0.9\textwidth]{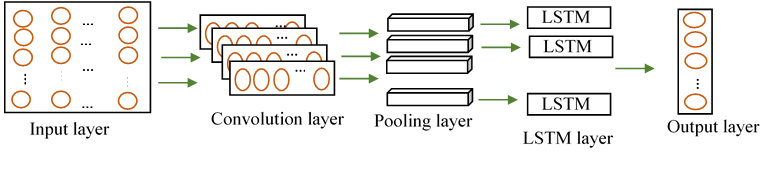}
\caption{Hybrid CNN--LSTM forecasting framework integrating convolutional
feature extraction and recurrent temporal modeling.}
\label{fig:hybrid}
\end{figure}

The forecaster pairs a convolutional feature extractor
(Figure~\ref{fig:cnn}) with a recurrent sequence model
(Figure~\ref{fig:lstm}), integrated into the hybrid pipeline of
Figure~\ref{fig:hybrid}. The CNN block applies $64$ one-dimensional
convolutional filters with kernel size $5$, followed by max-pooling, to
capture short-range fluctuations, promotional spikes, and day-of-week
effects. The resulting feature maps feed an LSTM block with $100$ hidden
units that captures medium- and long-horizon dependencies. The recurrent
dynamics follow the standard gated equations
\begin{align}
i_t&=\sigma(W_i x_t+U_i h_{t-1}+b_i), &
f_t&=\sigma(W_f x_t+U_f h_{t-1}+b_f),\\
\tilde c_t&=\tanh(W_c x_t+U_c h_{t-1}+b_c), &
o_t&=\sigma(W_o x_t+U_o h_{t-1}+b_o),\\
c_t&=f_t\odot c_{t-1}+i_t\odot\tilde c_t, &
h_t&=o_t\odot\tanh(c_t),
\end{align}
and a dense layer produces the point forecast $\hat y_t=W_y h_t+b_y$.
Training minimizes the mean squared error
\begin{equation}
\mathcal{L}=\frac{1}{N}\sum_{t\in T_{\text{train}}}(y_t-\hat y_t)^2 .
\end{equation}
Regularization uses dropout with rate $0.2$; optimization uses Adam with
initial learning rate $0.001$ and exponential decay factor $0.95$, with
early stopping (patience $20$ epochs) selected on the validation split. The
design rationale---combining a local-feature extractor with a sequence
model---follows established hybrid-architecture
evidence~\cite{Livieris2020,SeyvaniCNNFC2024,SeyvaniSubmatrix2025,GholizadehPINNDT2025}.

\subsubsection{Evaluation Metrics}
Forecast accuracy is assessed with scale-independent, time-series-appropriate
metrics: MAE, RMSE, MAPE, sMAPE, MASE (versus a na\"ive random walk), and
$R^{2}$:
\begin{align}
\text{MAE}&=\tfrac{1}{N}\!\sum_t|y_t-\hat y_t|, &
\text{RMSE}&=\sqrt{\tfrac{1}{N}\!\sum_t (y_t-\hat y_t)^2},\\
\text{MAPE}&=\tfrac{100}{N}\!\sum_t \tfrac{|y_t-\hat y_t|}{|y_t|}, &
\text{sMAPE}&=\tfrac{100}{N}\!\sum_t \tfrac{|y_t-\hat y_t|}{(|y_t|+|\hat y_t|)/2},\\
\text{MASE}&=\frac{\tfrac{1}{N}\sum_t|y_t-\hat y_t|}
{\tfrac{1}{N-1}\sum_t|y_t-y_{t-1}|}, &
R^{2}&=1-\frac{\sum_t (y_t-\hat y_t)^2}{\sum_t (y_t-\bar y)^2}.
\end{align}
A MASE below one indicates performance superior to the na\"ive baseline.

\subsection{Phase II: Multi-Objective Optimization Model}

\subsubsection{Modeling Assumptions}
The model is constructed under the following assumptions to ensure
analytical tractability and operational feasibility:
\begin{enumerate}[label=(A\arabic*),leftmargin=2.6em,itemsep=1pt]
\item Planning is multi-period and may be deterministic or scenario-based;
when uncertainty is included, scenario probabilities are known and stable.
\item Suppliers are classified as primary or backup; capacities and quality
indices are known at the start of the horizon.
\item All flows obey conservation of mass; no material is lost except
through controlled waste generation at consumption nodes.
\item Processing yields are constant across periods and technologies are
reliable.
\item Inventory ages accumulate linearly in discrete time, and freshness
decay follows a known monotonic function.
\item Market demand is known or scenario-dependent; customers are
non-strategic; unmet demand incurs shortage cost.
\item Waste-generation coefficients and circular conversion yields are
constant, and all collected waste is processable.
\item Transportation uses a finite set of modes with predefined
mode--route feasibility and constant lead times, costs, and emission
factors.
\item Facilities are operational throughout the horizon; plant activation
may change at the start of each period.
\item Products within a category are homogeneous; quality variability is
captured through yield and decay parameters.
\item Forward and reverse flows do not interfere except at circular
facilities and reintegration points.
\item Emission factors are known and constant; carbon pricing, if applied,
is uniform across time and scenarios.
\end{enumerate}

\subsubsection{Sets, Parameters, and Variables}
Index sets are $S=S^{P}\cup S^{B}$ (suppliers), $P$ (plants), $D$ (DCs),
$R$ (markets), $F$ (circular facilities), $K=K^{\text{fg}}\cup K^{\text{w}}$
(finished goods and waste), $M$ (transport modes), $T$ (periods), and
$\Omega$ (scenarios). Key parameters include unit
procurement/transport/processing/holding/shortage costs ($C^{\bullet}$),
mode emission factors ($E^{\bullet}$), processing yields $\mu_{p,k}$,
processing capacity $\text{Cap}^{\text{proc}}_{p,t}$, the freshness decay
rate $\theta_k$, the minimum acceptable freshness $\beta_k$, the maximum
allowable age $L_k^{\max}$, the target service level $\eta_{r,k}$, the
\emph{waste-generation rate} $\gamma_{r,k}$, the \emph{circular conversion
yield} $\rho_f$, circular capacity $\text{Cap}^{F}_{f,t}$, scenario
probabilities $\pi_\omega$, and the carbon price $P^{\mathrm{CO_2}}$.
Decision variables include forward/reverse flows $x^{\bullet}$, processing
$y_{p,k,t}^{\text{proc}}$, plant activation $u_{p,t}\in\{0,1\}$, inventories
$I^{P}_{p,k,t},I^{D}_{d,k,t}$, served demand $z_{r,k,t}$, shortage
$B_{r,k,t}$, service deviation $\sigma_{r,k,t}$, waste $w_{r,k,t}$, ages
$A^{P}_{p,k,t},A^{D}_{d,k,t}$, delivered freshness $q_{r,k,t}$, circular
returns $y^{F\to S}_{f,s,t},y^{F\to P}_{f,p,t}$, supplier-usage indicators
$u^{\text{sup}}_{s,p,t}\in\{0,1\}$, and mode-selection indicators
$y^{\text{mode}}_{i,j,k,t,m}\in\{0,1\}$. Throughout, $\gamma_{r,k}$ is the
\emph{only} symbol for the waste-generation rate and $\rho_f$ the \emph{only}
symbol for circular yield, removing the notation drift present in the thesis
draft.

\subsubsection{Objective Functions}
\textbf{Economic cost} aggregates procurement, transportation across all
arcs/modes, processing, inventory holding, shortage penalties, circular
collection/processing, and fixed plant-activation costs:
\begin{equation}
\begin{aligned}
\min Z^{\text{cost}}=
&\sum_{t,s,k}C^{\text{sup}}_{s,k}\!\sum_{p,m}x^{SP}_{s,p,k,t,m}
+\sum_{t,s,p,k,m}C^{SP}_{s,p,k,m}x^{SP}_{s,p,k,t,m}\\
&+\sum_{t,p,d,k,m}C^{PD}_{p,d,k,m}x^{PD}_{p,d,k,t,m}
+\sum_{t,d,r,k,m}C^{DR}_{d,r,k,m}x^{DR}_{d,r,k,t,m}\\
&+\sum_{t,p,k}C^{\text{proc}}_{p,k}y^{\text{proc}}_{p,k,t}
+\sum_{t,p,k}C^{\text{hold},P}_{p,k}I^{P}_{p,k,t}
+\sum_{t,d,k}C^{\text{hold},D}_{d,k}I^{D}_{d,k,t}\\
&+\sum_{t,r,k\in K^{\text{fg}}}C^{\text{short}}_{r,k}B_{r,k,t}
+\sum_{t,r,f}C^{\text{coll}}_{r,f}x^{RF}_{r,f,t}
+\sum_{t,f}C^{\text{circ}}_{f}\!\sum_{r}x^{RF}_{r,f,t}
+\sum_{t,p}C^{\text{fix}}_{p}u_{p,t}.
\end{aligned}
\end{equation}
\textbf{Carbon emissions} sum transportation and circular-processing
emissions:
\begin{equation}
\begin{aligned}
\min Z^{\text{em}}=\;
&\sum_{t,s,p,k,m}E^{SP}_{s,p,m}x^{SP}_{s,p,k,t,m}
+\sum_{t,p,d,k,m}E^{PD}_{p,d,m}x^{PD}_{p,d,k,t,m}\\
&+\sum_{t,d,r,k,m}E^{DR}_{d,r,m}x^{DR}_{d,r,k,t,m}
+\sum_{t,f}E^{\text{circ}}_{f}\!\sum_{r}x^{RF}_{r,f,t}.
\end{aligned}
\end{equation}
\textbf{Delivered freshness} is volume-weighted:
\begin{equation}
\max Z^{\text{fresh}}=\sum_{t,r,k\in K^{\text{fg}}}q_{r,k,t}\,z_{r,k,t}.
\end{equation}

\subsubsection{Constraints}
\textbf{Material balance.} Inbound supply feeds production through the yield
coefficient, and conservation holds at plants, DCs, and markets:
\begin{align}
\sum_{s,m}x^{SP}_{s,p,k,t,m}&=\frac{y^{\text{proc}}_{p,k,t}}{\mu_{p,k}},
&&\forall p,k,t,\\
\sum_{d,m}x^{PD}_{p,d,k,t,m}&=y^{\text{proc}}_{p,k,t}+I^{P}_{p,k,t-1}-I^{P}_{p,k,t},
&&\forall p,k,t,\\
\sum_{p,m}x^{PD}_{p,d,k,t,m}+I^{D}_{d,k,t-1}&=\sum_{r,m}x^{DR}_{d,r,k,t,m}+I^{D}_{d,k,t},
&&\forall d,k,t,\\
\sum_{d,m}x^{DR}_{d,r,k,t,m}&=z_{r,k,t}+w_{r,k,t},
&&\forall r,k\in K^{\text{fg}},t.
\end{align}
\textbf{Production and capacity.} Production respects yield, plant capacity,
and activation:
\begin{align}
y^{\text{proc}}_{p,k,t}\le \mu_{p,k}\!\sum_{s,m}x^{SP}_{s,p,k,t,m},\qquad
\sum_{k}y^{\text{proc}}_{p,k,t}\le \text{Cap}^{\text{proc}}_{p,t}\,u_{p,t},\qquad
y^{\text{proc}}_{p,k,t}\le M^{\text{proc}}u_{p,t}.
\end{align}
\textbf{Inventory dynamics} at plants and DCs, with non-negativity:
\begin{align}
I^{P}_{p,k,t}&=I^{P}_{p,k,t-1}+\sum_{s,m}x^{SP}_{s,p,k,t,m}
-y^{\text{proc}}_{p,k,t}-\sum_{d,m}x^{PD}_{p,d,k,t,m}, &&\forall p,k,t,\\
I^{D}_{d,k,t}&=I^{D}_{d,k,t-1}+\sum_{p,m}x^{PD}_{p,d,k,t,m}
-\sum_{r,m}x^{DR}_{d,r,k,t,m}, &&\forall d,k,t,\\
I^{P}_{p,k,t}&\ge 0,\quad I^{D}_{d,k,t}\ge 0. &&
\end{align}
\textbf{Demand fulfilment and service level.} Served demand is bounded,
shortage is recorded, and a minimum service fraction is enforced with a
deviation variable:
\begin{align}
z_{r,k,t}=\sum_{d,m}x^{DR}_{d,r,k,t,m},\quad
z_{r,k,t}\le D_{r,k,t},\quad
B_{r,k,t}=D_{r,k,t}-z_{r,k,t},\quad
z_{r,k,t}\ge \eta_{r,k}D_{r,k,t}-\sigma_{r,k,t}.
\end{align}
\textbf{Freshness.} Age accumulates at plants and DCs, and delivered
freshness decays exponentially and must meet the minimum threshold and
maximum-age limit:
\begin{align}
A^{P}_{p,k,t}&=\frac{I^{P}_{p,k,t-1}\,(A^{P}_{p,k,t-1}+1)}{I^{P}_{p,k,t}},\quad
A^{D}_{d,k,t}=\frac{I^{D}_{d,k,t-1}\,(A^{D}_{d,k,t-1}+1)}{I^{D}_{d,k,t}},\\
q_{r,k,t}&=\exp\!\big(-\theta_k A^{D}_{d,k,t}\big),\qquad
q_{r,k,t}\ge\beta_k,\qquad A^{D}_{d,k,t}\le L_k^{\max}.
\end{align}
\textbf{Circular flows} use a single waste-rate symbol $\gamma_{r,k}$ and a
single yield symbol $\rho_f$, with circular capacity:
\begin{gather}
w_{r,k,t}=\gamma_{r,k}\,z_{r,k,t},\qquad
\sum_{f}x^{RF}_{r,f,t}=w_{r,k,t},\qquad
\sum_{r}x^{RF}_{r,f,t}\le\text{Cap}^{F}_{f,t},\notag\\
\sum_{s}y^{F\to S}_{f,s,t}+\sum_{p}y^{F\to P}_{f,p,t}=\rho_f\!\sum_{r}x^{RF}_{r,f,t}.
\end{gather}
\textbf{Sourcing resilience and mode selection.} The primary/backup mix is
bounded, supplier diversification is enforced, and each shipment selects at
most one feasible mode:
\begin{align}
\sum_{s\in S^{P}}\!\sum_{p,k,m}x^{SP}_{s,p,k,t,m}&\ge\delta\!\!\sum_{s\in S}\!\sum_{p,k,m}x^{SP}_{s,p,k,t,m},\quad
\sum_{s\in S^{B}}\!\sum_{p,k,m}x^{SP}_{s,p,k,t,m}\le\phi\!\!\sum_{s\in S}\!\sum_{p,k,m}x^{SP}_{s,p,k,t,m},\\
\sum_{s}u^{\text{sup}}_{s,p,t}&\ge\kappa_p,\qquad
\sum_{m}y^{\text{mode}}_{i,j,k,t,m}\le 1,\qquad
x_{i,j,k,t,m}\le M^{\text{big}}\,y^{\text{mode}}_{i,j,k,t,m}.
\end{align}
\textbf{Forecast linkage.} The demand parameter is fixed to the ML output,
in deterministic or scenario form, with normalized probabilities:
\begin{equation}
D_{r,k,t}=D^{\text{ML}}_{r,k,t},\qquad
D^{\omega}_{r,k,t}=D^{\text{ML}}_{r,k,t}(\omega),\qquad
\sum_{\omega\in\Omega}\pi_\omega=1,\;\pi_\omega\ge 0.
\end{equation}
Standard non-negativity and binary-domain restrictions apply to all
remaining variables.

\subsubsection{Solution Approach}
The tri-objective model is solved with the $\varepsilon$-constraint method:
cost is minimized while emissions and freshness are bounded,
\begin{equation}
\min Z^{\text{cost}}\quad\text{s.t.}\quad
Z^{\text{em}}\le\varepsilon^{\text{em}},\;\;
Z^{\text{fresh}}\ge\varepsilon^{\text{fresh}},
\end{equation}
and the bounds are swept across their feasible ranges (obtained from the
single-objective optima) to trace the Pareto frontier. For comparative
visualization and a complementary weighted-sum analysis, the objectives are
normalized to a common dimensionless scale,
\begin{equation}
\tilde Z^{\text{cost}}=\frac{Z^{\text{cost}}-Z^{\text{cost},\min}}{Z^{\text{cost},\max}-Z^{\text{cost},\min}},\quad
\tilde Z^{\text{em}}=\frac{Z^{\text{em}}-Z^{\text{em},\min}}{Z^{\text{em},\max}-Z^{\text{em},\min}},\quad
\tilde Z^{\text{fresh}}=\frac{Z^{\text{fresh},\max}-Z^{\text{fresh}}}{Z^{\text{fresh},\max}-Z^{\text{fresh},\min}},
\end{equation}
and aggregated as
$Z^{\text{WS}}=\lambda^{\text{cost}}\tilde Z^{\text{cost}}
+\lambda^{\text{em}}\tilde Z^{\text{em}}
+\lambda^{\text{fresh}}\tilde Z^{\text{fresh}}$.
The model is implemented in Python; the MILP is solved with Gurobi and the
forecaster is built in TensorFlow/Keras. The planning horizon is $|T|=12$
periods throughout.

\section{Results and Discussion}\label{sec:results}

\subsection{Data Description}
\begin{figure}[H]
\centering
\includegraphics[width=0.72\textwidth]{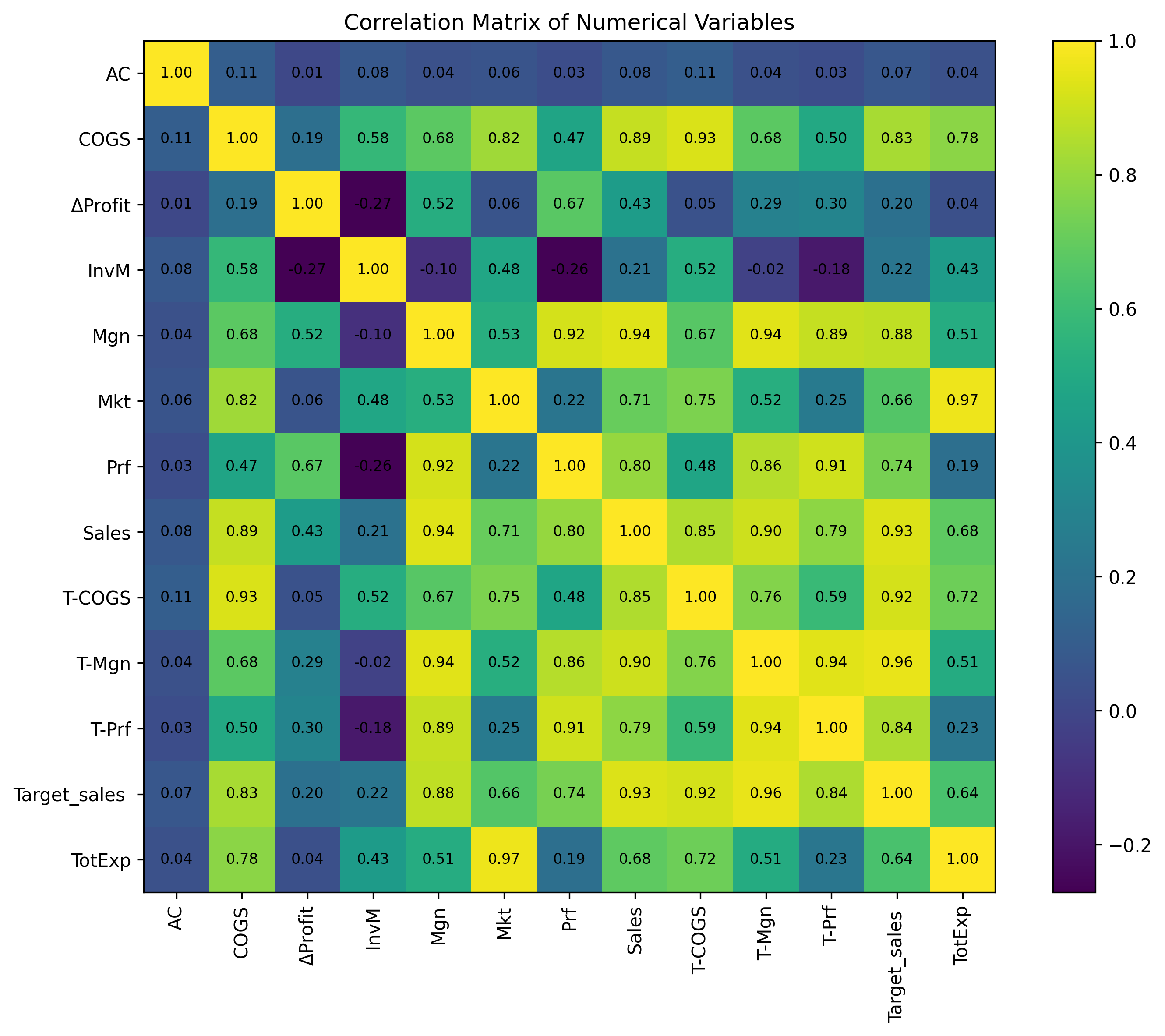}
\caption{Correlation matrix of numerical variables.}
\label{fig:corr}
\end{figure}

\begin{figure}[H]
\centering
\includegraphics[width=0.82\textwidth]{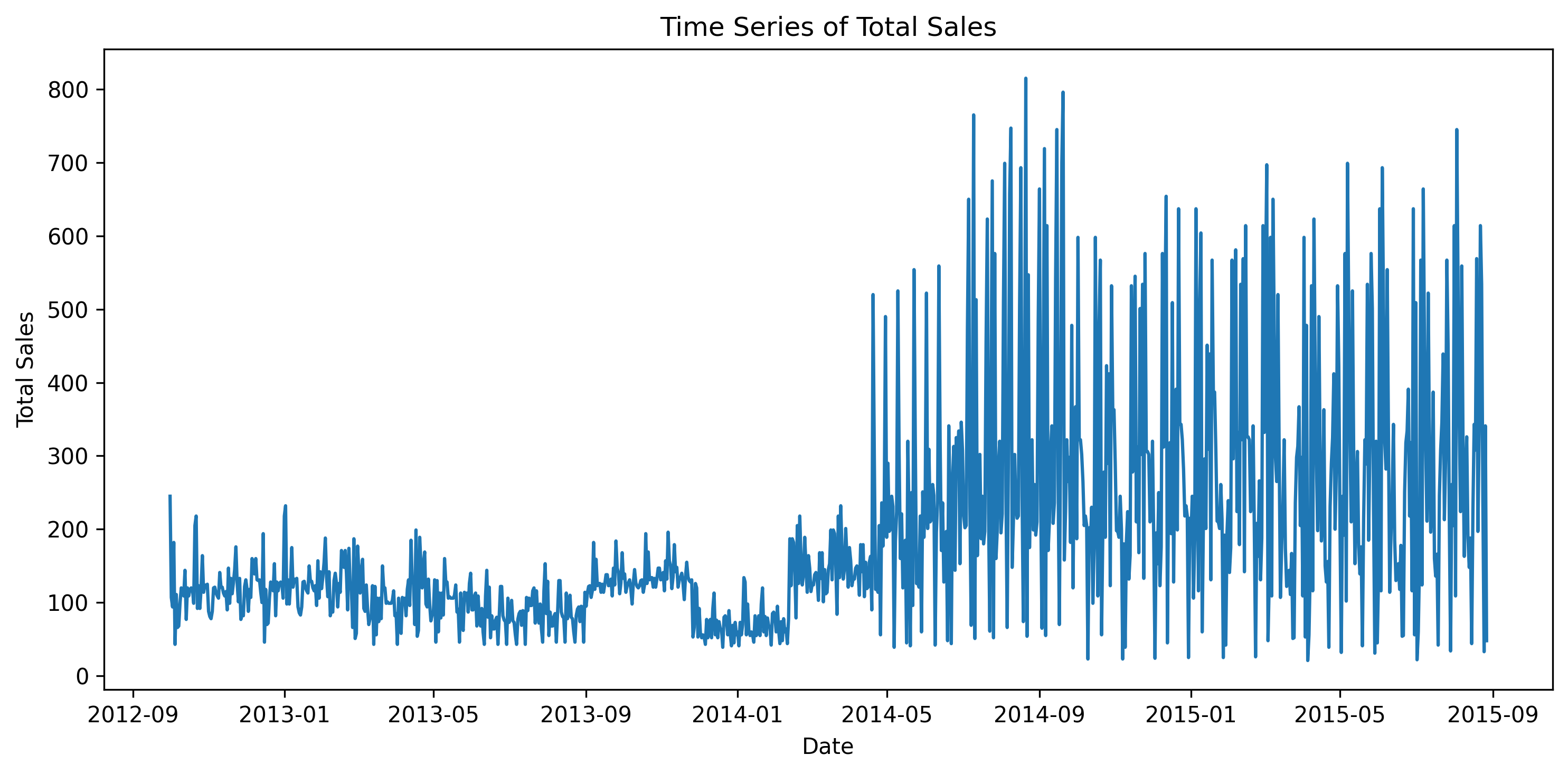}
\caption{Time series of total sales over the study period.}
\label{fig:timeseries}
\end{figure}

\begin{figure}[H]
\centering
\includegraphics[width=0.82\textwidth]{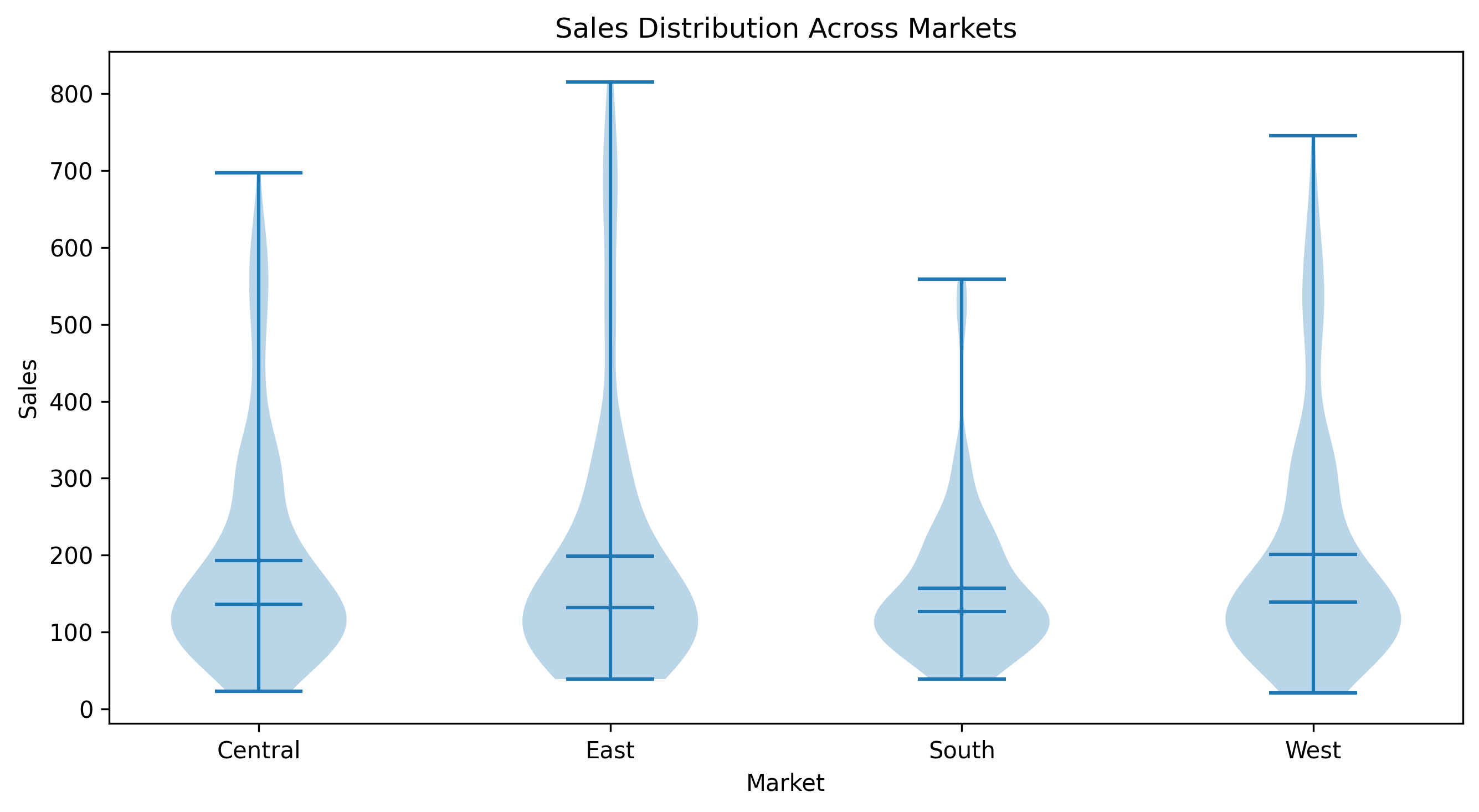}
\caption{Distribution of sales across markets (violin plots).}
\label{fig:violin}
\end{figure}

The selected dataset reflects real demand behavior across multiple products
and U.S. markets. Table~\ref{tab:descstats} reports descriptive statistics
of the main numerical variables, and Table~\ref{tab:cv} reports
coefficients of variation (CV). The relatively high CVs for profit,
inventory margin, and sales indicate substantial variability, reinforcing
the need for nonlinear, data-driven forecasting. Figures~\ref{fig:corr},
\ref{fig:timeseries}, and~\ref{fig:violin} show, respectively, the
correlation structure, the temporal evolution of aggregate sales, and the
heterogeneity of sales across markets.

\begin{table}[H]
\centering
\caption{Descriptive statistics of numerical variables in the Coffee Chain
Sales dataset.}
\label{tab:descstats}
\small
\setlength{\tabcolsep}{4pt}
\begin{tabular}{lrrrrrr}
\toprule
\textbf{Variable} & \textbf{Mean} & \textbf{Std.\ dev.} & \textbf{Median} &
\textbf{Mode} & \textbf{Min} & \textbf{Max}\\
\midrule
Area code        & 587.03 & 225.30 & 573  & 435 & 203  & 970\\
Cogs             & 82.40  & 64.82  & 57   & 54  & 17   & 308\\
Inventory margin & 815.18 & 916.16 & 659  & 601 & 10   & 3{,}980\\
Margin           & 102.42 & 91.29  & 73   & 71  & 15   & 427\\
Marketing        & 30.43  & 25.96  & 22   & 14  & 0    & 100\\
Profit           & 60.56  & 100.52 & 39.5 & 27  & $-181$ & 778\\
Sales            & 191.05 & 148.27 & 133  & 120 & 31   & 760\\
Target margin    & 96.82  & 89.47  & 70   & 60  & 15   & 427\\
Target profit    & 60.17  & 77.82  & 40   & 30  & $-50$ & 300\\
Target sales     & 168.49 & 145.96 & 120  & 90  & 31   & 760\\
Total expenses   & 53.84  & 31.70  & 46   & 46  & 11   & 156\\
\bottomrule
\end{tabular}
\end{table}

\begin{table}[H]
\centering
\caption{Variability indicators of key numerical variables.}
\label{tab:cv}
\begin{tabular}{lrrr}
\toprule
\textbf{Variable} & \textbf{Mean} & \textbf{Std.\ dev.} & \textbf{CV}\\
\midrule
Sales            & 191.05 & 148.27 & 0.78\\
Marketing        & 30.43  & 25.96  & 0.85\\
Margin           & 102.42 & 91.29  & 0.89\\
Profit           & 60.56  & 100.52 & 1.66\\
Cogs             & 82.40  & 64.82  & 0.79\\
Inventory margin & 815.18 & 916.16 & 1.12\\
Total expenses   & 53.84  & 31.70  & 0.59\\
\bottomrule
\end{tabular}
\end{table}

\subsection{Forecasting Performance}
\begin{figure}[H]
\centering
\includegraphics[width=0.72\textwidth]{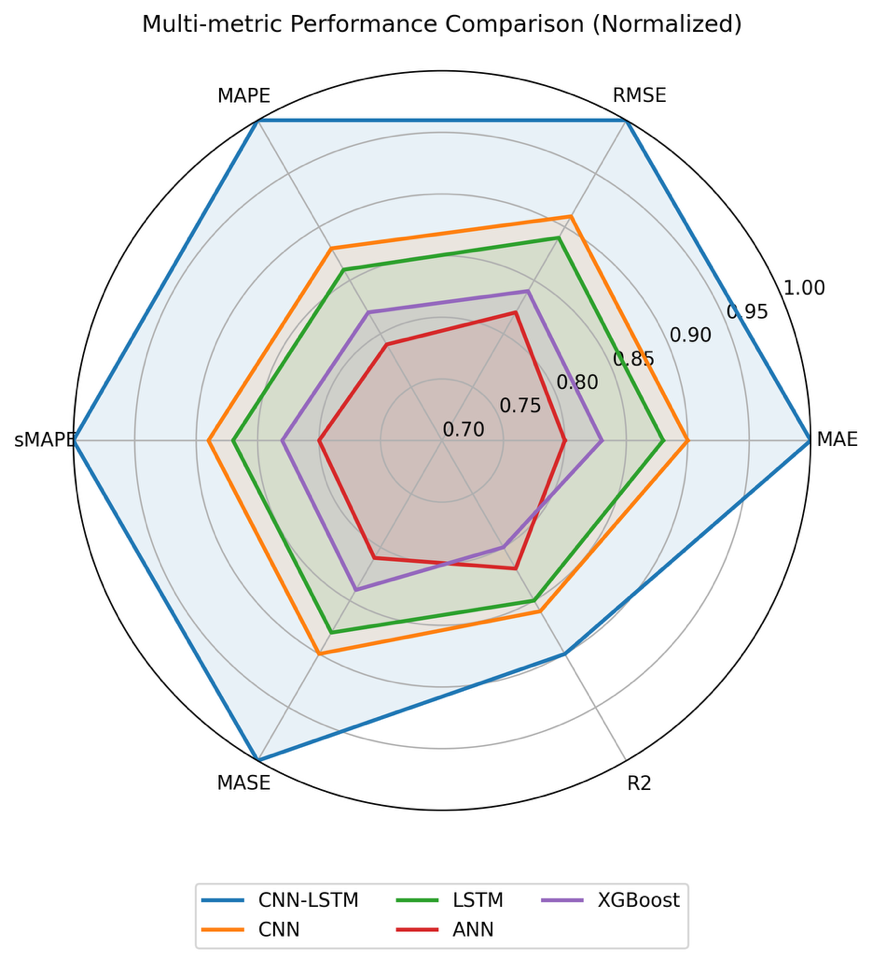}
\caption{Normalized multi-metric performance comparison.}
\label{fig:multimetric}
\end{figure}

Table~\ref{tab:forecast} compares the hybrid CNN--LSTM against ten
benchmarks under the chronological 70/15/15 protocol. The hybrid model is
best on every metric, achieving MAE $=22.87$, RMSE $=32.94$, and
$R^{2}=0.90$. It improves MAE by $\approx12\%$ over standalone CNN and
$\approx11.8\%$ over standalone LSTM, and by more than $30\%$ over classical
statistical baselines---consistent with hybrid-architecture evidence in
other temporal and mapping
domains~\cite{Livieris2020,SeyvaniCNNFC2024,NikrouTransferable2025}.
Figure~\ref{fig:multimetric} summarizes the multi-metric comparison.

\begin{table}[H]
\centering
\caption{Forecasting performance comparison across models (chronological
70/15/15 split).}
\label{tab:forecast}
\begin{tabular}{lrrrrrr}
\toprule
\textbf{Model} & \textbf{MAE} & \textbf{RMSE} & \textbf{MAPE (\%)} &
\textbf{sMAPE (\%)} & \textbf{MASE} & \textbf{$R^{2}$}\\
\midrule
Na\"ive (random walk) & 41.32 & 58.47 & 24.85 & 22.91 & 1.00 & 0.62\\
Moving average        & 36.78 & 51.26 & 21.43 & 20.17 & 0.89 & 0.68\\
ARIMA                 & 34.15 & 48.92 & 19.84 & 18.65 & 0.83 & 0.71\\
SVR                   & 31.74 & 45.18 & 18.02 & 17.21 & 0.76 & 0.75\\
Random forest         & 29.63 & 42.06 & 16.47 & 15.88 & 0.71 & 0.78\\
XGBoost               & 28.41 & 40.92 & 15.96 & 15.21 & 0.69 & 0.80\\
ANN (MLP)             & 27.85 & 39.73 & 15.42 & 14.88 & 0.67 & 0.82\\
LSTM                  & 25.94 & 36.88 & 14.21 & 13.65 & 0.62 & 0.85\\
CNN                   & 25.31 & 36.12 & 13.98 & 13.42 & 0.61 & 0.86\\
\textbf{Hybrid CNN--LSTM} & \textbf{22.87} & \textbf{32.94} &
\textbf{12.36} & \textbf{11.98} & \textbf{0.55} & \textbf{0.90}\\
\bottomrule
\end{tabular}
\end{table}

A Diebold--Mariano (DM) test (Table~\ref{tab:dm}) confirms statistically
significant improvements over all classical and traditional-ML benchmarks
at the $5\%$ level; gains over the strongest standalone deep models reach
marginal significance ($p=0.069$ for LSTM, $p=0.091$ for CNN), reflecting
the incremental nature of improvement over already-strong base learners.

\begin{table}[H]
\centering
\caption{Statistical significance of forecasting improvements
(Diebold--Mariano test).}
\label{tab:dm}
\begin{tabular}{lrrrc}
\toprule
\textbf{Compared model} & \textbf{Mean error diff.} & \textbf{DM stat.} &
\textbf{$p$-value} & \textbf{Significance}\\
\midrule
Na\"ive          & $-18.45$ & $-4.82$ & $<0.001$ & Significant\\
Moving average   & $-13.91$ & $-3.97$ & $<0.001$ & Significant\\
ARIMA            & $-11.28$ & $-3.45$ & $0.001$  & Significant\\
SVR              & $-8.87$  & $-2.94$ & $0.004$  & Significant\\
Random forest    & $-6.76$  & $-2.41$ & $0.016$  & Significant\\
XGBoost          & $-5.54$  & $-2.08$ & $0.038$  & Significant\\
ANN (MLP)        & $-4.98$  & $-1.97$ & $0.049$  & Significant\\
LSTM             & $-3.07$  & $-1.82$ & $0.069$  & Marginal\\
CNN              & $-2.44$  & $-1.69$ & $0.091$  & Marginal\\
\bottomrule
\end{tabular}
\end{table}

\subsection{Optimization Results}
With forecast demand as input, the $\varepsilon$-constraint method is run on
a \emph{simulated but industrially-consistent} numerical instance: total
demand of $2{,}067$ units over $|T|=12$ periods, three market nodes, and two
finished products (espresso, filter), with a calibrated shortage penalty of
$70.74$ per unit. The $\varepsilon$ ranges are $[522.63,\,1{,}075.86]$ for
emissions and $[1{,}834.63,\,2{,}067]$ for freshness, derived from the
single-objective optima. Table~\ref{tab:opt} reports the selected compromise
solution: full demand satisfaction (zero shortage), binding emissions at
$799.25$, and maximal freshness ($2{,}067$).

\begin{table}[H]
\centering
\caption{Optimal objective values for the selected compromise solution
(simulated instance).}
\label{tab:opt}
\begin{tabular}{lr}
\toprule
\textbf{Objective} & \textbf{Optimal value}\\
\midrule
Total cost $Z^{\text{cost}}$        & 13{,}058.30\\
Carbon emissions $Z^{\text{em}}$    & 799.25\\
Total freshness $Z^{\text{fresh}}$  & 2{,}067.00\\
\bottomrule
\end{tabular}
\end{table}

\subsection{Trade-off Analysis}
\begin{figure}[H]
\centering
\includegraphics[width=0.88\textwidth]{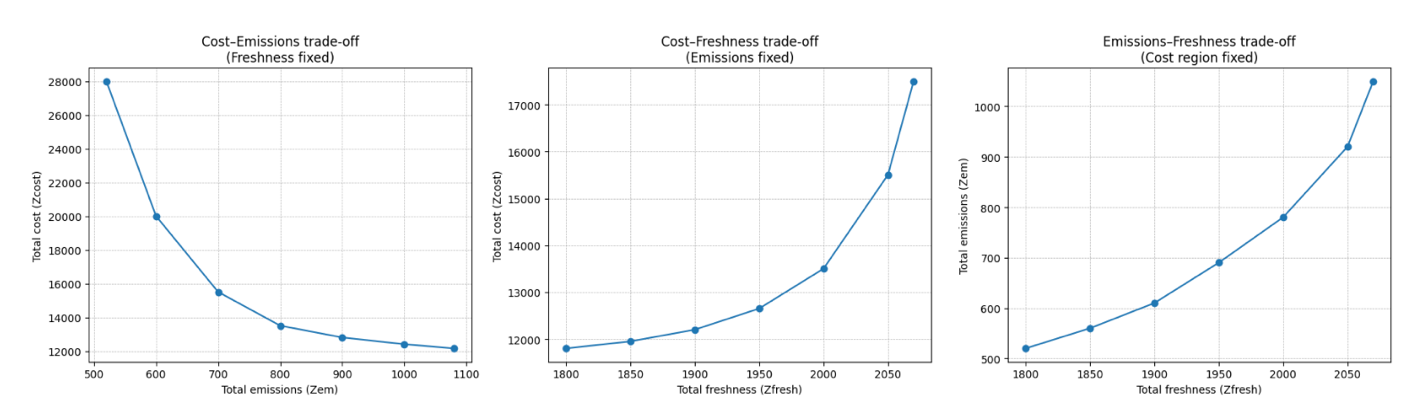}
\caption{Pairwise trade-offs among cost, emissions, and freshness.}
\label{fig:pareto}
\end{figure}

Sweeping the $\varepsilon$ bounds yields $25$ Pareto-efficient solutions.
Figure~\ref{fig:pareto} shows the pairwise trade-offs. The cost--emission
curve is strongly nonlinear: reducing emissions from $\approx1{,}050$ to
$\approx800$ raises cost modestly (from $\approx13{,}000$ to
$\approx14{,}500$), whereas pushing below $\approx600$ drives cost above
$28{,}000$---a threshold effect arising once cheap abatement options are
exhausted. Higher freshness generally requires faster, higher-emission
transport, producing an emission--freshness conflict.

\subsection{Single-Parameter Sensitivity Analysis}
\begin{figure}[H]
\centering
\includegraphics[width=0.80\textwidth]{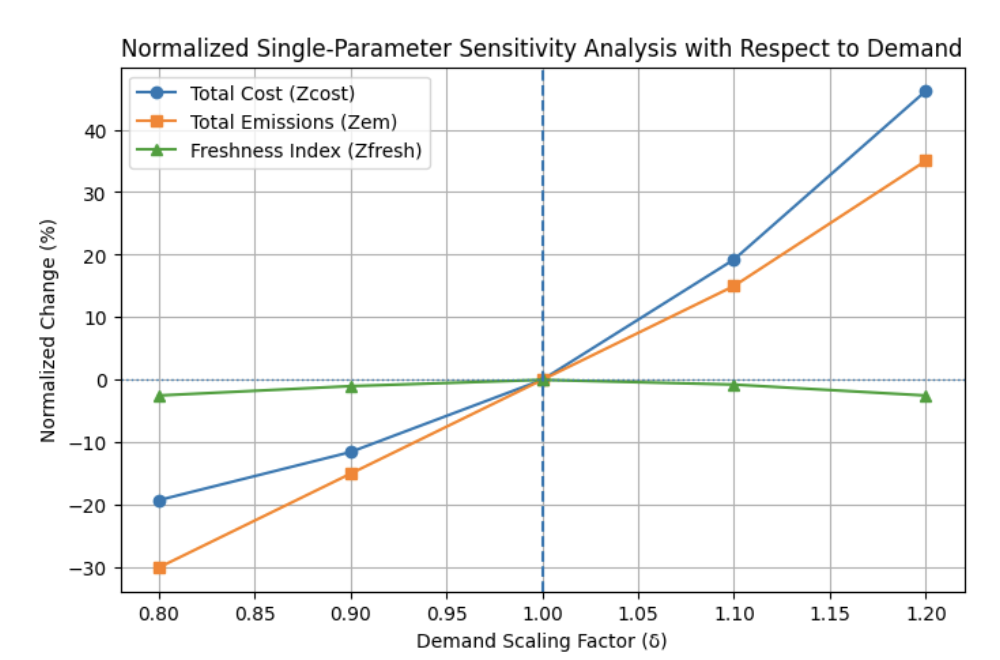}
\caption{Normalized sensitivity to the demand scaling factor $\delta$.}
\label{fig:sens-demand}
\end{figure}

\begin{figure}[H]
\centering
\includegraphics[width=0.80\textwidth]{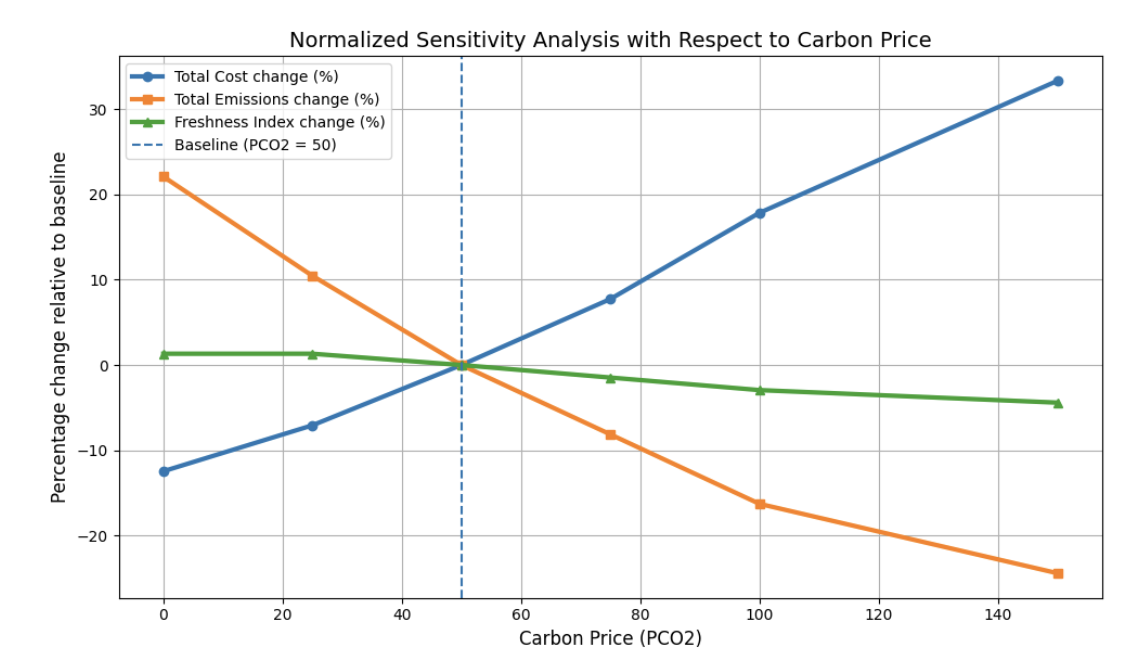}
\caption{Normalized sensitivity to the carbon price $P^{\mathrm{CO_2}}$.}
\label{fig:sens-carbon}
\end{figure}

\begin{figure}[H]
\centering
\includegraphics[width=0.80\textwidth]{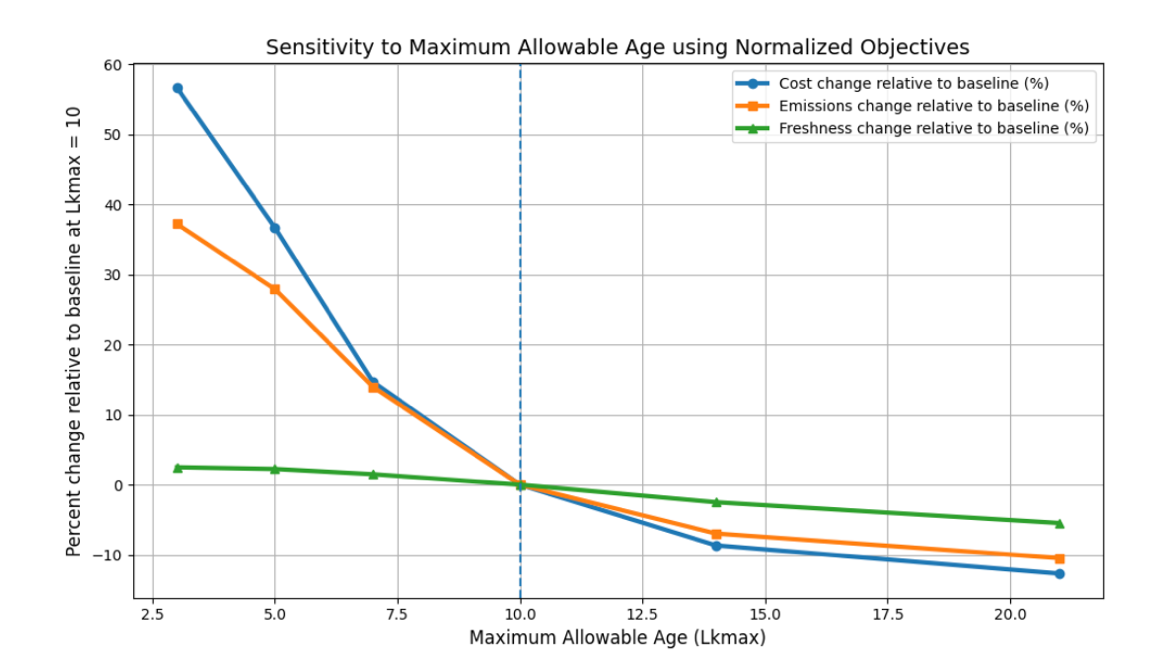}
\caption{Normalized sensitivity to the maximum allowable product age
$L_k^{\max}$.}
\label{fig:sens-age}
\end{figure}

Single-parameter analyses
(Figures~\ref{fig:sens-demand}--\ref{fig:sens-age}) vary the demand scaling
factor $\delta\in[0.8,1.2]$, the carbon price
$P^{\mathrm{CO_2}}\in[0,150]$, and the maximum allowable age $L_k^{\max}$.
Cost rises monotonically and nonlinearly with demand---accelerating at high
$\delta$ as backup suppliers and premium modes activate---while freshness is
non-monotonic, peaking near baseline turnover and declining at extremes.
Carbon pricing monotonically lowers emissions and raises cost; freshness is
stable at low--moderate prices and declines at high prices as slower,
low-emission modes dominate. Strict age limits ($L_k^{\max}\le 7$) impose
cost and emission premiums exceeding $30$--$50\%$. These abrupt,
constraint-driven regime shifts mirror the threshold behavior reported for
data-driven decision support in other
settings~\cite{GholizadehThreshold2024,NikrouDecision2025}.

\subsection{Bivariate Sensitivity Analysis}
\begin{figure}[H]
\centering
\includegraphics[width=0.92\textwidth]{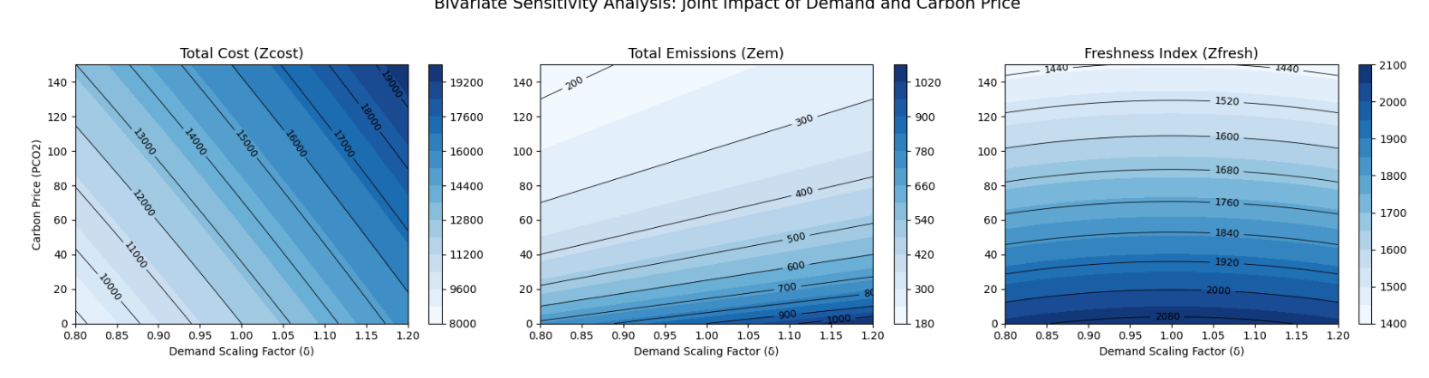}
\caption{Joint sensitivity of objectives to demand ($\delta$) and carbon
price.}
\label{fig:joint-dc}
\end{figure}

\begin{figure}[H]
\centering
\includegraphics[width=0.92\textwidth]{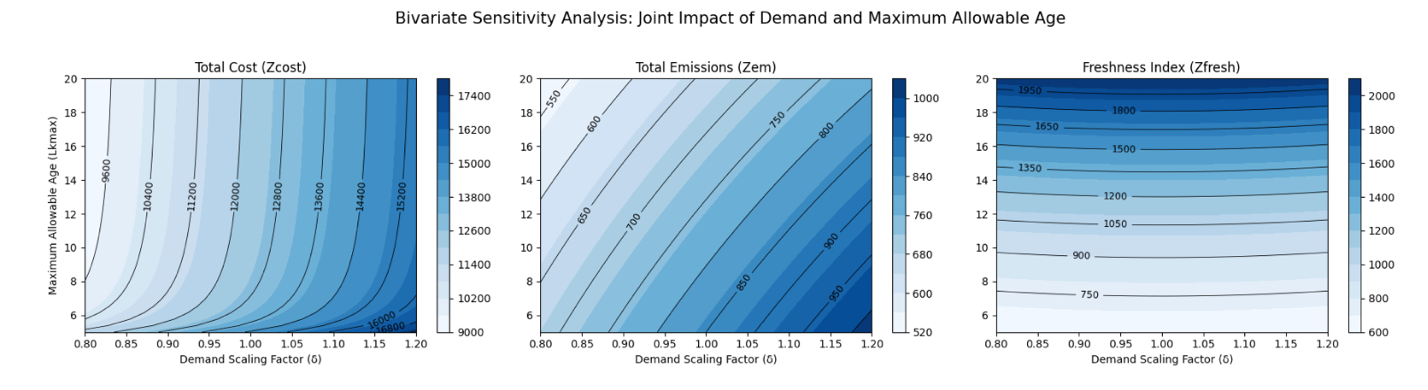}
\caption{Joint sensitivity of objectives to demand ($\delta$) and maximum
allowable age.}
\label{fig:joint-da}
\end{figure}

\begin{figure}[H]
\centering
\includegraphics[width=0.92\textwidth]{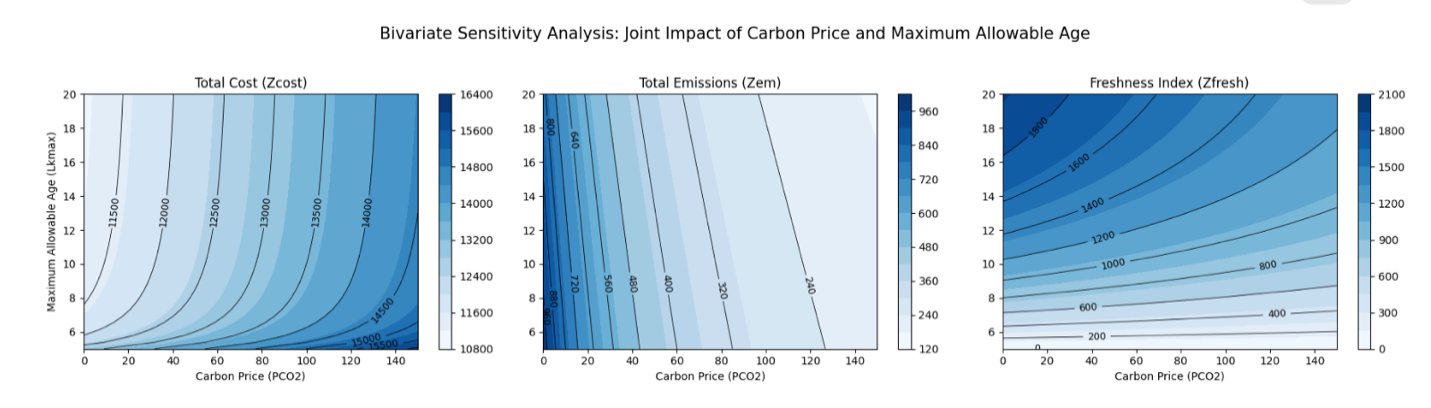}
\caption{Joint sensitivity of objectives to carbon price and maximum
allowable age.}
\label{fig:joint-ca}
\end{figure}

Bivariate analyses vary parameter pairs simultaneously to expose
interaction effects (Figures~\ref{fig:joint-dc}--\ref{fig:joint-ca}). The
joint demand--carbon-price surface shows a reinforcing interaction in cost
(the economic burden of carbon policy grows with market pressure) while
carbon pricing suppresses emissions across all demand levels. The joint
demand--age surface shows densely packed cost contours under strict age
limits, indicating abrupt cost escalation when high demand meets tight
freshness requirements. The joint carbon-price--age surface confirms that
high freshness is attainable only when age limits are generous and carbon
prices are low, highlighting an indirect but significant impact of
environmental policy on delivered quality.

\subsection{Operational Indicators and Policy-Scenario Analysis}
\begin{figure}[H]
\centering
\includegraphics[width=0.74\textwidth]{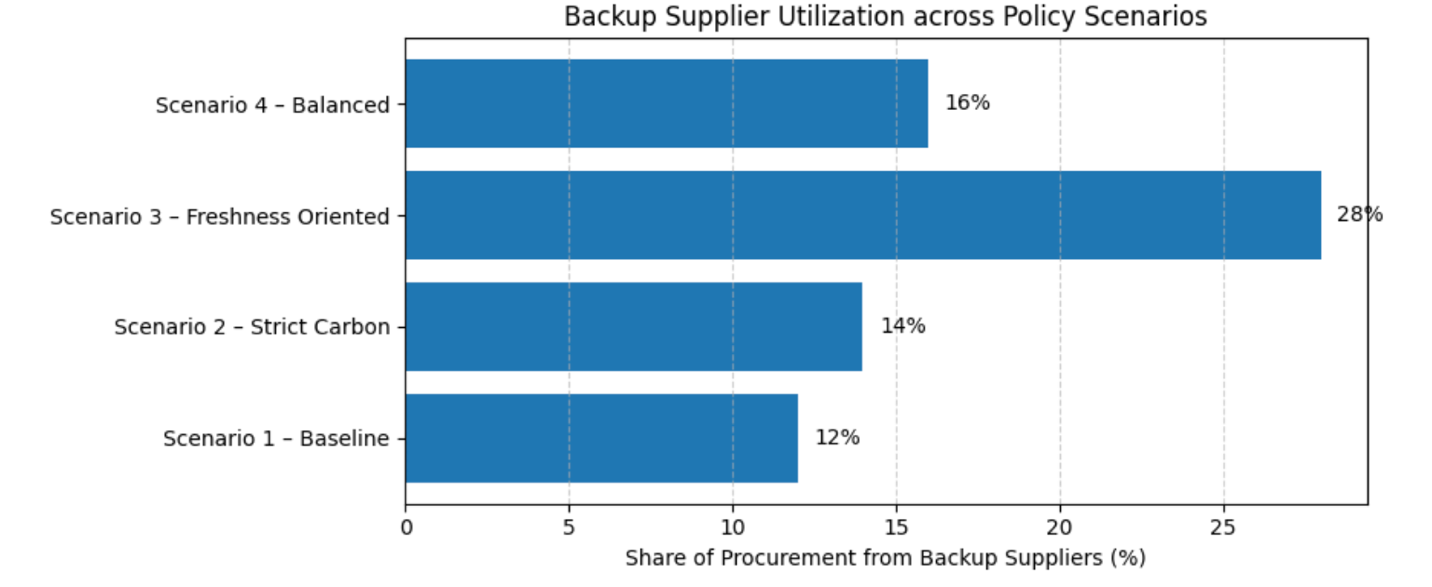}
\caption{Share of procurement from backup suppliers across policy
scenarios.}
\label{fig:backup}
\end{figure}

Beyond the three objective values, Table~\ref{tab:kpi} reports operational
indicators for the selected compromise solution, providing a comprehensive
view of how the network operates under the optimal configuration. Four
scenarios then represent distinct managerial orientations: (1) baseline,
(2) strict carbon, (3) freshness-oriented, and (4) balanced sustainability.
Table~\ref{tab:policy} compares objectives and operational indicators, and
Figure~\ref{fig:backup} shows backup-supplier reliance. The balanced policy
delivers the most favorable multi-dimensional outcome: a $9.9\%$ cost
increase buys a $22.4\%$ emission reduction while freshness stays within
$1.7\%$ of baseline---clear evidence that coordinated, moderate policies
outperform extreme single-objective strategies. Modal shifts are the
mechanism: strict carbon pushes sea share to $50\%$ and air to $3\%$,
whereas the freshness-oriented policy raises air to $22\%$ and cuts sea to
$18\%$.

\begin{table}[H]
\centering
\caption{Key performance indicators of the selected compromise solution.}
\label{tab:kpi}
\begin{tabular}{lrl}
\toprule
\textbf{Indicator} & \textbf{Value} & \textbf{Unit}\\
\midrule
Total demand fulfilled          & 2{,}067 & units\\
Service level                   & 100     & \%\\
Average product age at delivery & 8.0     & periods\\
Road transport share            & 54      & \%\\
Maritime transport share        & 36      & \%\\
Air transport share             & 10      & \%\\
Primary supplier share          & 84      & \%\\
Backup supplier share           & 16      & \%\\
Circular recovery rate          & 65      & \%\\
Active processing plants        & 2       & facilities\\
\bottomrule
\end{tabular}
\end{table}

\begin{table}[H]
\centering
\caption{Objective values and operational indicators across policy
scenarios (simulated instance).}
\label{tab:policy}
\begin{tabular}{lrrrr}
\toprule
\textbf{Indicator} & \textbf{Baseline} & \textbf{Strict carbon} &
\textbf{Freshness} & \textbf{Balanced}\\
\midrule
Total cost $Z^{\text{cost}}$      & 12{,}850 & 14{,}980 & 17{,}640 & 14{,}120\\
\;\; vs.\ baseline                & ---      & $+16.6\%$ & $+37.3\%$ & $+9.9\%$\\
Total emissions $Z^{\text{em}}$   & 980      & 610      & 1{,}120  & 760\\
\;\; vs.\ baseline                & ---      & $-37.8\%$ & $+14.3\%$ & $-22.4\%$\\
Freshness index $Z^{\text{fresh}}$& 2{,}040  & 1{,}925  & 2{,}085  & 2{,}005\\
\;\; vs.\ baseline                & ---      & $-5.6\%$  & $+2.2\%$  & $-1.7\%$\\
\midrule
Air transport share               & 8\%  & 3\%  & 22\% & 10\%\\
Road transport share              & 52\% & 47\% & 60\% & 54\%\\
Sea transport share               & 40\% & 50\% & 18\% & 36\%\\
Avg.\ product age at delivery      & 9.2  & 11.1 & 5.4  & 8.0\\
Service shortage rate             & 2.5\% & 3.1\% & 1.2\% & 2.0\%\\
Recycling / collection rate       & 62\% & 68\% & 55\% & 65\%\\
Backup-supplier share             & 12\% & 14\% & 28\% & 16\%\\
\bottomrule
\end{tabular}
\end{table}

\section{Conclusions}\label{sec:conc}

\subsection{Summary and Contributions}
This study developed an integrated two-phase framework that links
deep-learning demand forecasting to multi-objective optimization for
circular coffee supply chains. The hybrid CNN--LSTM forecaster (MAE
$22.87$, $R^{2}=0.90$ under a strict chronological 70/15/15 protocol)
supplies time-indexed demand directly to a tri-objective MILP that
simultaneously minimizes cost, minimizes emissions, and maximizes delivered
freshness, with freshness modeled as exponential decay of inventory age.
The $\varepsilon$-constraint method produced $25$ Pareto-efficient
solutions, and sensitivity and policy analyses revealed pronounced
threshold effects. The main innovations are: explicit integration of
forecasting with optimization; treatment of freshness as an independent
objective with dynamic age tracking; a nonlinear exponential decay
formulation of freshness; a closed-loop model centered on coffee waste;
simultaneous modeling of multimodal transport, freshness, and emissions;
backup suppliers as a resilience mechanism; structured trade-off analysis
via the $\varepsilon$-constraint method; multi-dimensional sensitivity and
policy analyses; and an integrated decision-support framework for balanced
policy design.

\subsection{Managerial Implications}
Effective coffee-chain management cannot be driven by isolated objectives;
decisions focused exclusively on cost, emissions, or quality produce
structurally imbalanced outcomes. Demand forecasting should be treated as a
core tactical input rather than a background activity, because realistic,
time-dependent demand patterns align production, inventory, and distribution
with actual market behavior. Freshness should be managed as a time-based
coordination problem via explicit age tracking, since freshness outcomes are
highly sensitive to storage, mode selection, and turnover. Environmental
policies such as carbon pricing reshape transportation and sourcing with
indirect effects on cost and quality that managers must anticipate. Flexible
sourcing structures---with backup suppliers used as strategic buffers rather
than routine sources---improve resilience. Finally, balanced policies
consistently dominate extreme single-objective strategies.

\subsection{Limitations and Future Research}
Several limitations should be acknowledged. The forecasting data are public
and may not fully represent every operational context; the Phase-II instance
is a simulated (industrially-consistent) one; several parameters are
deterministic and time-invariant; and strategic facility-location decisions
are out of scope. Future work will incorporate stochastic or robust
optimization to address uncertainty; integrate endogenous consumer-choice
models linking freshness, price, and demand; expand to strategic network
design and capacity expansion; explore alternative or interactive
multi-objective methods; develop richer circular-economy process models; and
conduct empirical case studies across regions. Building on advances in
IoT/digital-twin sensing, blockchain-secured data layers, and human-centered
AI adoption~\cite{GholizadehIoT2025,GholizadehPINNDT2025,EsmaeiliAICS2024,MirzaeiBIM2025},
a particularly promising direction is the coupling of real-time traceability
data with dynamic optimization as these technologies mature.


\end{document}